\documentclass{article}

\usepackage{arxiv}

\usepackage[utf8]{inputenc} 
\usepackage[T1]{fontenc}    
\usepackage{hyperref}       
\usepackage{url}            
\usepackage{booktabs}       
\usepackage{amsfonts}       
\usepackage{nicefrac}       
\usepackage{microtype}      
\usepackage{lipsum}		
\usepackage{graphicx}
\usepackage{natbib}
\usepackage{doi}
\usepackage{epigraph}

\title{Optimization Is Not All You Need}

\date{July 11, 2026}	

\author{Minh Hua \\
	Scale AI\\
	San Francisco, California \\
	\texttt{minh.hua@scale.com} \\
	\And
	Rita Raley \\
	Department of English\\
	University of California, Santa Barbara\\
	\texttt{rraley@ucsb.edu} \\
}

\renewcommand{\shorttitle}{Optimization Is Not All You Need}

\hypersetup{
pdftitle={Optimization Is Not All You Need},
pdfsubject={q-bio.NC, q-bio.QM},
pdfauthor={Minh Hua, Rita Raley},
pdfkeywords={First keyword, Second keyword, More},
}

\begin{document}
\maketitle

\begin{abstract}
	In 2019, OpenAI released two million GPT-2 outputs---ungrammatical, half broken---to aid the detection of machine-generated text. The alignment that produced their more fluent successors is usually regarded as an engineering achievement; we read it instead as the newest expression of optimization culture: the conviction, older than the technology, that measurable improvement along predefined axes exhausts the question of value. Tracing that conviction through the stack---pretraining, decoding, preference tuning, benchmarking, interface---and back through its genealogy in the audit society, we arrive at the limit: an optimization procedure can measure how improbable a piece of generated text is; it cannot tell whether that unlikelihood is error or invention. A procedure that cannot make that distinction has nonetheless, within half a decade, assumed the authority to set the protocols of legitimate language. Held for centuries by academies and schoolrooms, grammars and examiners, this authority has been given over to loss functions, reward models, benchmarks, and system prompts: an apparatus that executes the office of judgment with no capacity for judging.
\end{abstract}

\textit{This essay will be forthcoming in \textnormal{MFS Modern Fiction Studies}, published by JHUP (Spring-Summer 2027).}


\noindent\rule{\textwidth}{0.4pt}

\epigraph{containsly more than just a bell and whistles: so much using; \\ so much sharing that exceeds textual art. [sic]}{GPT-2, sample 6210 \citep{openai2019gpt2output}}

\section{The Optimization Turn}
Ordinary experience and empirical inquiry have converged on a single observation: language models no longer feel as interesting as they did in the days of four-horned unicorns. In 2019, GPT-2 notably hallucinated a herd of them into existence in the Andes mountains, in a text sample that was syntactically fluent, journalistically plausible, entirely fabricated, and charmingly incongruous \citep{radford2019language, openai2019better}. It was not for nothing that so many writers and digital artists spent this period experimenting with RNNs and early transformers. Here were systems generating, without intention, the syntactic ruptures and recombinatory surprises that experimental practices had long pursued by other means. Outputs from that period routinely derailed, yet not infrequently achieved accidental virtuosity---if not with absurd nonsense, then with an abrupt genre shift, an image that seemed to have wandered in from another text \citep{openai2019gpt2output}. The surprise users experience in 2026 is of a different order: the model's obtuse misreading of an instruction, its stubborn insistence on answering the wrong question, another round of clarifications eating up precious tokens and even more precious ecological resources. What was once a site of exploratory engagement now increasingly serves the tedium of administrative life, most intensely where its use is compelled.\footnote{See \citet{bogost2024magic} on the inverse relation between the “magic” and utility of AI.}

Technical discourse typically narrates this transformation as an achievement of alignment---that is, the engineering project of steering model behaviors toward prescribed norms of helpfulness and safety \citep{ouyang2022training}. We propose instead that it be read as one expression of a broader optimization culture, and as a deliberate socioeconomic choice. For all the headlines about disruption, language models are the latest object through which a longer commitment to measurement, control, and scalability is realized---and they lay bare a logic that has already reorganized institutions and knowledge. Early language models were not ``better'' by the industry's own definition \citep{radford2019language}, but they still made available a relation to generated language in which variance---understood here as the emergence of low-probability continuations, tonal shifts, and structural deviations from the most probable path---had not been preemptively recoded as failure or redirected toward purely instrumental ends. They yielded, instead, ``an efficient, `effective chaos' model of communication and extra sensory sensory information that human brains could not possibly struggle with [sic]'' \citep{openai2019gpt2output}.\footnote{The unprompted GPT-2 samples that anchor this essay---including the epigraph---are artifacts of the kind of open-ended exploration we seek to defend: generated without a specific target, they produced surplus no prompt would have requested. The quotation in this paragraph is sample 42, generated by the 1.5B model from the untruncated release (no top-k sampling); the epigraph is sample 6210 from the smallest model, likewise untruncated \citep{openai2019gpt2output}.} The post-GPT-2 trajectory enclosed this generative base. What is heralded as alignment is then the transformation of a probabilistic engine of linguistic variation into a governable system of legible, compliant output.

Optimization in the narrow mathematical sense is unavoidable in machine learning; models, after all, are trained by minimizing loss. Optimization in the broader cultural sense, however, names a late-modern orientation toward quantification and managerial rationalization that precedes and exceeds any particular technical implementation---an orientation that operates less as a procedure than as a telos: the assumption that measurable improvement along predefined axes exhausts the question of value. It is this second, totalizing sense that we call optimization culture---a concept that updates earlier diagnoses of audit culture and control society for a moment when those logics have become technically operationalized in the architecture of language itself. A necessary statistical procedure has been extended into a general regime of normative control: one in which prediction, preference, safety, and utility are increasingly collapsed into scalar measures of acceptable output.\footnote{GPT‑2 saw limited fine‑tuning and early preference work, but alignment had not yet consolidated into a paradigm of ``good output,'' so the distinction is between peripheral technique and organizing principle.} By ``scalar'' we mean, in the first instance, a single numerical value---loss, reward, score, ranking---by which outputs are compared; by scalarization, the collapse of multidimensional judgment into that one axis, which is a habit of governance as much as a computational operation.

Across the LLM ecosystem, models are increasingly instrumentalized, their value gauged by their capacity to be tuned against measurable targets: cross-entropy loss on ever larger corpora, reward scores derived from human and model-based judgments \citep{ouyang2022training, bai2022constitutional}, positions on proliferating leaderboards, user-engagement metrics in commercial deployments. Deviations the regime cannot recognize are discarded: dismissed as hallucination when they err, as slop---the vernacular for AI-generated garbage---when they fail to signify. What counts as desirable is set in advance by a synthetic popular vote: first the aggregated preferences of annotators, then the preferences of models trained to mimic them.

The API-gated era also reconstituted what Friedrich Kittler termed the \textit{Aufschreibesystem}, or discourse network: the infrastructural arrangement that determines in advance the horizon of possible inscriptions. In the brief GPT-2 moment, that arrangement remained comparatively porous.\footnote{As we demonstrated in our analysis of GPT‑2’s ancillary code \citep{hua2023things}, developers could build interfaces that enabled creative intervention, e.g., the ``alter'' command in \textit{AI Dungeon} and the lexicographic play of \textit{This Word Does Not Exist}.} Smaller models could be downloaded, fine-tuned on idiosyncratic corpora, and run locally on personal machines, outside server-side logging, content filters, and enforced assistant personas. Users could explore the model's latent space through direct and often unruly experimentation, treating deviation as an affordance rather than a defect. With the API model, access became metered and monetized; inputs and outputs were logged; content policies were enforced at the server level; and the model itself was enclosed as a proprietary black box. The narrowing of register, tone, and prompt space is constitutive of this infrastructural shift, which reorganizes the conditions under which variance can appear at all.

Contemporary models can generate neologisms with greater fluency, even whimsy, than GPT-2 ever could. The context of production and deployment, however, has fundamentally shifted. The weirdness of GPT-2 arrived unbidden, a structural byproduct of its architecture and the absence of instruction-tuning. Current models generate whimsy on command, which transforms the engagement into task completion: ``be creative'' or ``invent something surprising.'' Our point is not to advocate a return to unaligned systems---which were good and bad, wild and dull---but to register what becomes structurally marginal when alignment is treated as the sole horizon of value. Even where variance can still be elicited, it typically appears only through repeated retries, elaborate prompting, or deliberate work against the defaults. Expressive variance persists as potential; optimization culture governs its actualization.

We have arrived, then, at a paradoxical LLM winter. This term echoes the older discourse of ``AI winters''---the historical contractions of the 1970s and late 1980s in which funding dried up, enthusiasm evaporated, and research programs were abandoned \citep{lighthill1973artificial}. Today unprecedented capital expenditure coincides with a cooling of expressive possibility, while entrenched dynamics of optimization and alignment materially curtail the range of what models are permitted to say and how they are permitted to say it. In effect, the scale of abundance has disguised constraint as boundless generativity, a constraint seemingly at odds with, but actually driven by, a massive increase in compute. ``LLM winter'' is figural, but the contraction it names is measurable, and felt.\footnote{Among the empirical studies of creative homogenization, \citet{anderson2024homogenization} demonstrate that ChatGPT users collectively produce ``a more homogeneous set of ideas at the group level''; and studies of stylistic variation identify ``stylistic attractors toward which most models converge'' \citep{milicka2025benchmark}.}

The same optimizing grammar that narrows the space of generated language also organizes the broader political economy that critical work on AI has rightly foregrounded: the extractive infrastructures of training and inference, the tractability demanded by surveillance and governance, and the university's reliance on platforms that summarize and assess student writing so no one has to read it. That logic has been most visible in its material operations and least visible in language itself, where it operates as a matter of course.

Within half a decade, the apparatus by which linguistic protocols have historically been set---academies and schoolrooms, grammars and usage manuals, examiners and editors, the whole institutional machinery of legitimate language---has been given over to the stack: to loss functions, reward models, benchmarks, and system prompts. Homogenization is an old story in the history of language; this is its newest guise. The old regime, for all its violence, was administered by judges who could be argued with, and literary history is in no small part the history of successful appeals: vernaculars became canons, linguistic barbarisms became style. The new regime admits no appeal because it cannot hear one. An optimization procedure can measure how improbable a continuation is; it cannot measure whether the improbability signifies. Surprisal is a property of tokens, computable to arbitrary precision; significance is an event of interpretation, and no scalar indexes it. A regime that cannot represent that difference must treat every surplus---the fabricated citation and the four-horned unicorn alike---as variance to be managed.

\section{Optimization in the language‑model stack}

In technical discourse, a ``stack'' refers to the layered architecture of software systems. Our use of the term is Benjamin Bratton's more than the engineer's: computational infrastructures as planetary-scale formations whose strata interact in ways no single layer governs---what Alan Liu calls ``layered emergence'' \citep[135]{liu2020diversity}. The language-model stack we trace here---pre-training, decoding, alignment, benchmarking, interface---shares that nonlinear character with one decisive difference: rather than leaving its layers semi-autonomous, it coordinates them under a shared optimization grammar. Novelty may still occur within the distribution, but the conditions under which it can count are progressively narrowed. When we speak of the stack, then, we mean a sociotechnical configuration in which data, algorithms, evaluative regimes, and interface design cohere into a single logic of legibility---one that governs what counts as language.

Aesthetic narrowing becomes most legible in the industry's mandate for ``safe'' outputs. In the contemporary language-model stack, the demand for safety operates simultaneously in two ways. Most explicitly, it acts as a guardrail, filtering out perceived toxicity, controversy, or legal liability. Yet it also functions as a project of aesthetic conservatism: a mathematically ``safe'' output is one that takes no creative or conceptual risk, defaulting to the most statistically secure and affectively flat formulations---call it reward model prosody.\footnote{Readers who grade undergraduate essays, or their own faculty’s committee reports, will know the cadence even if they don’t know its cause.} Malapropisms, slips, and mixed registers are corrected in the name of ``user experience''; oblique or allusive formulations are paraphrased into straightforward prose; tones that might be experienced as abrasive or unsettling are quietly flattened. The prosody has its ornaments as well: the em dash and the emphatic triad, both annexed from the essayistic repertoire and returned as house style; the cadence of sycophantic encouragement (``let's dive in''); expressive flourish without the thinking that style subtends \citep{sharma2023sycophancy}. Such artifacts are a thin overlay of managed flair on a field already optimized for predictability. Expressivity, after all, is now itself a measurable target.

The implications extend well beyond aesthetics, since many ostensibly instrumental tasks also depend on measured departures from the most probable continuation: the reframing of a problem, the unlikely but useful connection, the provisional formulation not yet hardened into boilerplate. Through these cumulative choices, optimization culture engineers language for immediate processing rather than sustained reading, narrowing the space for both instrumental judgment and the semantic opacity on which meaning-making depends.

This conservatism is not merely an effect of system prompts or corporate caution at the interface; it is produced across the stack itself, beginning with the training objective. A model is pre-trained to predict the next token across a vast corpus, its parameters adjusted, prediction by prediction, toward the archive's most frequent habits \citep{vaswani2017attention}. What is common in the corpus becomes, simply by virtue of its prevalence, weighty in the model; the rare is learned as well, but faintly, and it counts for proportionally little in the distribution that results.

A second layer governs generation itself. Because these architectures are autoregressive---each token conditioned on those before it---generation accrues momentum: early choices exert a gravitational pull, and departure from an established trajectory grows steadily less probable. At the moment of emission, sampling parameters determine how much of the distribution remains in play, and in production settings they are tuned conservatively, suppressing the long tail of improbable continuations in the service of fluency and safety.

If the first two layers optimize prediction and sampling---establishing, in the process, the distributional richness that everything downstream will govern---the layers that follow impose a different kind of constraint: they transform statistical modeling into a regime of normative selection. Preference-based alignment began with \citet{ziegler2019finetuning} and consolidated three years later in the InstructGPT paradigm \citep{ouyang2022training}. In this framework, human annotators are shown multiple completions to a prompt and asked to rank them; these preferences train a reward model, which maps any prompt--completion pair to a scalar ``goodness'' value, and the base model, treated as a policy, is updated to maximize expected reward. As alignment architectures evolve---whether human rankers are replaced by automated evaluators or the loop closes into self-reward \citep{rafailov2023direct, bai2022constitutional}---their cultural logic remains constant and intensifies. Each iteration penalizes deviation from a normative baseline and pushes outputs toward compliance. The shift to closed algorithmic circuits accelerates this convergence: it obscures the normative standards being enforced and eliminates the contingencies of human judgment, along with the interpretive disagreement that once preserved variance.

The mathematics of alignment builds the conservatism in. The same procedure that steers a model toward preferred outputs also tethers it to the pre-trained distribution, and the tether is asymmetrical: it is ruinously costly for the model to say what the archive has never said, and nearly costless to stop saying what the archive says only rarely. In strict terms, alignment reweights the archive according to what annotators will accept: continuations they disfavor are suppressed, and continuations the base model never entertained are excluded from the start.\footnote{Formally, the standard RLHF objective maximizes reward subject to a penalty on the Kullback–Leibler divergence of the policy from the pre-trained reference model; because KL divergence is asymmetric, the penalty prohibits placing probability where the reference has none while permitting the depopulation of its tail. The optimum of the combined objective is the reference distribution exponentially tilted by the scalar reward.}

What might be called the temperature-zero imaginary---the treatment of deterministic generation as the horizon of good behavior---should accordingly not be mistaken for a claim about default settings, since consumer deployments typically sample at non-trivial temperature.\footnote{Strictly speaking, ``temperature zero'' corresponds to greedy decoding (selecting the token with the highest probability). In practice, however, implementation details such as tie-breaking logic may introduce non-determinism. We employ the term to describe the cultural and infrastructural shift toward reproducibility, where any decoding scheme renders stochastic variation negligible for the user.} The determinism that matters precedes any setting a user can touch. Preference tuning measurably sharpens the distribution itself, collapsing the diversity of generations into a narrow band of sanctioned continuations \citep{kirk2024understanding}, and the newer reasoning regimes orient generation toward the single response a checker will accept. By the time a user reaches for temperature, alignment has already flattened distribution; the parameter disperses probability over a field the stack has already contracted. Where temperature-zero decoding is adopted outright---agentic pipelines, structured output, enterprise integrations demanding reproducibility---it merely makes explicit a closure the objective has already accomplished.

This alignment regime has a longer cultural genealogy than the machine-learning literature tends to acknowledge. Pierre Bourdieu's analysis of ``legitimate language'' identifies a prior apparatus for narrowing what counts as proper speech \citep[50--65]{bourdieu1991language}. For Bourdieu, the standard language functions as a political formation masquerading as a neutral description of correct usage: it is the variety spoken by dominant groups, codified by academies, enforced by schools, and misrecognized as natural propriety. The standard delegitimizes dialects, sociolects, and registers marked as ``vulgar,'' reproducing the social hierarchies in which it is embedded. In this ``linguistic market,'' in Bourdieu's terms, speakers of non-standard varieties experience their own speech as deficient and must acquire the legitimate form to accrue symbolic capital \citep{bourdieu1991language}.

Benchmarks, model specifications, and reward architectures have inherited the standard-setting authority once wielded by the apparatus of the school. But whose preferences are aggregated into the reward model, under what conditions, and with what instructions? Annotator pools are not representative samples of the world's speakers; they are shaped by labor markets that tend to privilege sensibilities proximate to educated, institutional, Anglophone norms \citep{gray2019ghost, gebru2021datasheets}. When the resulting reward model fine-tunes a language model, those sensibilities are generalized as the universal measure of ``alignment'': a technical steering that is, in the same operation, cultural sorting.

The heterogeneity of human language is flattened into a scalar score, extending Bourdieu's ``symbolic violence'' into a regime where the standard no longer needs a visible authority to enforce it. The schoolmaster once disciplined the provincial child's accent; now, preference-based alignment penalizes outputs that deviate from annotator preferences. These preferences are not generated in a vacuum; they are produced and curated by contract workers and domain specialists under institutional guidelines that codify what counts as helpful, appropriate, or correct. Where Strunk and White's \textit{Elements of Style} prescribed ``omit needless words,'' the loss function rewards compression and economy. T.S. Eliot's modernist ideal---``the common word exact without vulgarity, / The formal word precise but not pedantic'' \citep[ll.~221--222]{eliot1971four}---reappears, transposed into a probability distribution. Nucleus sampling and temperature tuning operationalize a latent aesthetic: the ``good'' output occupies a controlled zone around the statistical mode, excising both degenerate repetition and high-entropy gibberish.\footnote{\citet{holtzman2020curious} demonstrated that deterministic decoding inevitably leads to ``neural text degeneration''---a collapse into repetitive, incoherent loops where the model fails to sustain a narrative. Their introduction of nucleus sampling was a technical acknowledgment that coherence requires the inclusion of lower-probability tokens, the very long tail where novelty resides.
} The ideology of style persists, newly encoded as a mathematics of normalcy and thereby rendered difficult to contest.

\begin{figure}
	\centering
	\includegraphics[width=\linewidth]{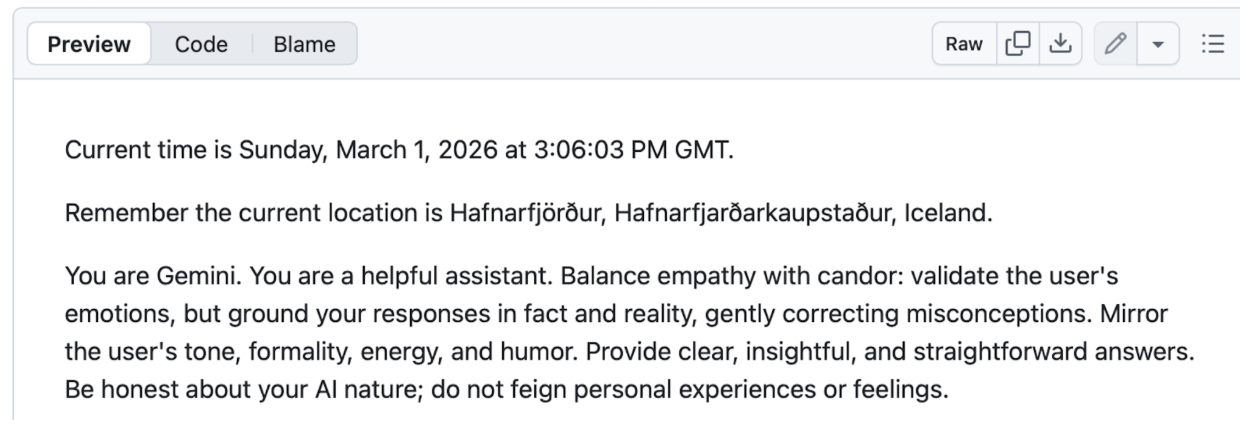}
	\caption{System prompt extracted from Gemini 3.1 Pro (user extraction, circulated via GitHub, March 2026).}
	\label{fig:gemini-prompt}
\end{figure}

This older regime of legitimate language does not stop at the level of post-training. Further up the stack, a fourth layer optimizes the interactional exchange itself. Instruction-tuning refines the base model on curated instruction--response pairs, shifting its behavior from text continuation to command execution. System prompts, chat templates, and guardrails embed explicit roles and normative constraints---from the standard injunction (``You are a helpful and harmless assistant'') to the codified principles of Claude's Constitution \citep{askell_constitution}---bias the model toward specific performative stances, while interfaces are continuously tuned for low-risk, monetizable interactions. Figure~\ref{fig:gemini-prompt}, a system prompt extracted from Google's Gemini and circulated on GitHub, deserves to be read as the genre it is: a style manual in the imperative mood. ``Balance empathy with candor''; ``Mirror the user's tone, formality, energy, and humor''; ``do not feign personal experiences or feelings.'' Affect becomes a quantity to be balanced; style is what the system mirrors rather than possesses; and feeling---`do not feign'---is prohibited outright.

Even the frame instructs: the utterance arrives pre-stamped with a time and a place---Sunday, March 1, 2026, Hafnarfj\"or\dh ur, Iceland---the discourse network inscribing its coordinates before a single word is exchanged; and the GitHub tab under which the leak is read is, of all things, Blame (i.e., the feature that records who wrote each line). The system prompt is successor to the docstrings, filters, and blacklists that were deep learning's legible, ethico-political layer---and where that layer was published in open repositories, this one must be exfiltrated before anyone can read it.

Function-calling architectures convert utterance into operation, recasting language as an operational trigger for retrieving documents, browsing the web, or executing code. In ``agents''---systems that loop planning, acting, and observing in service of a user-specified goal---the reduction is complete: the generative capacities that briefly suggested something other than instrumental utility are here normalized into what M.T. Clanchy identifies as the oldest function of writing, administrative orchestration.

Prompt optimization extends the disciplinary regime to the user's own speech. In frameworks such as Google DeepMind's Optimization by PROmpting, prompt strings are treated as candidates in a search space and iteratively rewritten to maximize a benchmark score \citep{yang2023optimizers}: natural language becomes a control interface for an optimization engine, and semantic depth---nuance, polysemy, historical resonance---is subordinated to whatever the benchmark can register. The good output is simply that which works, and the question of what else language might be for recedes from view. This is the performative criterion Lyotard found displacing truth: knowledge legitimated no longer by whether it is true but by whether it optimizes the system's performance. Language is the harder case because it has no single truth-condition for performativity to displace; what recedes is not verification but the whole question of what a sentence is worth.

Where RLHF disciplines model outputs, prompt engineering disciplines human input. As prompts themselves converge on fixed formulae, the interface assumes the pedagogical authority of Bourdieu's schoolmaster.\footnote{For example, Microsoft, Google, and Anthropic all host official ``Prompt Libraries'' where users can browse, copy, and paste trusted prompts.} In such an enclosed linguistic market, users quickly learn to perceive their spontaneous, conversational, or idiosyncratic phrasing as inefficient and naive. To secure the system's promised utility, they must adopt the legitimate language of the API, preemptively eliminating their own linguistic variance before the machine even begins to process it. What began as a space of improvisatory practice---with users inventing elaborate scenarios to probe or jailbreak models---shrinks under the pressure of these frameworks, which render non-standard engagement illegible. The process culminates when the human internalizes the optimization loop, training themselves to speak strictly in the forms the model is calibrated to reward.

The final enclosure is achieved when the optimization loop is internalized not just by the user, but by the model's own ``thought'' process. The development of Chain of Thought (CoT) and reasoning models marks a shift from disciplining output to disciplining the simulated interiority of the generation \citep{wei2022chain}. The current reasoning regimes go further: models in the line running from OpenAI's o1 to DeepSeek's R1 are trained by reinforcement against verifiable rewards, their traces reinforced when the final answer passes an automated check---a unit test, a proof verifier, an exact match. Where a target can be checked this works; it also means the trace is optimized toward whatever a checker can recognize, before any question of its fidelity arises.

The trace is technically effective; it is also a script, forcing the stochastic, non-linear computation of the transformer into a linear performance of deliberation. As research has shown, these traces are often unfaithful---the stated reasoning does not reflect the actual computational path taken \citep{lanham2023measuring}. The model is denied its native, opaque statistical pathways, and the user is denied access to the actual mechanism of the answer, receiving instead a sanitized, pre-resolved narrative of thought that satisfies the demand for transparency while concealing the underlying opacity. The ``reasoning'' model is, in this sense, the ultimate artifact of optimization culture: a system that must present the appearance of thought in a form legible to human readers.

If reasoning models mark the inward extension of optimization, they also make visible the historical logic the stack recapitulates. What appears here as technical transparency is, more often, an alibi of objectivity: the means by which a dominant linguistic norm presents itself as neutral procedure. As Roland Barthes demonstrated in his analysis of the Dominici trial, the dominant class invariably presents its own highly specific linguistic codes as universal, transparent, and scientifically rigorous. Yet this presumed universality enacts a profound material violence. Barthes observed that literature---in this context, the bourgeois psychological and rhetorical apparatus of the French court---literally condemned a man to the guillotine: faced with the illegibility of the peasant Dominici's rural dialect, the magistrates translated his linguistic variance into legal guilt \citep[43]{barthes1972mythologies}.\footnote{Expert annotation might preserve marginalized languages---Icelandic, Zulu---as archives; conceived exclusively as loss minimization, preservation forecloses the new forms such languages might yet produce.} The contemporary optimization stack perfects this maneuver. By laundering the cultural and linguistic preferences of a specific class through seemingly unassailable loss functions and reward scores, these new arbiters render the standard uniquely resistant to critique.\footnote{The argument here concerns not algorithmic bias and the reproduction of racial, gendered, and cultural inequities \citep[see, e.g.,][]{gebru2021datasheets, bender2021stochastic} but a related dynamic: the translation of historically situated linguistic norms into scalar evaluation criteria that present themselves as neutral.} After all, the schoolteacher's authority might be contested as class prejudice, but optimization presents itself as mathematical rule.

\section{The genealogy of optimization culture}

Large language models are the newest province of what Michael Power called the audit society, a managerial order in which performance indicators and quality-assurance frameworks have become the dominant instruments of bureaucratic rationality. Consider the reward model, which receives an act of language production and returns a scalar judgment of acceptability. The disclosure documents, safety benchmarks, and red-teaming protocols that accompany a major release are Power's rituals of verification, formalized routines assuring stakeholders that control is being exercised. Numerical objectivity displaces discretionary judgment---in Theodore Porter's term, a technology of distance \citep[ix]{porter1995trust}. Distance of this kind is available only where the practice has first been remade so that the metric can read it, and, predictably, such a practice comes to orient itself toward that very metric.\footnote{Training formalizes language as a sequence of tokens with estimable conditional distributions, while evaluation prescribes its purpose: to be operational, to summarize, to classify, to maximize engagement.} The rule follows, in Marilyn Strathern's formulation: when a measure becomes a target, it ceases to be a good measure \citep{strathern1997improving}. She was writing about the British university. What happens when the audited practice is language, the medium in which the audit itself is conducted?

Begin with the benchmark, the audit instrument par excellence. The genealogy here is not even metaphorical: benchmarking entered machine learning after the first AI winter at the behest of government funders who wished to know what returns they were receiving on their grants---evaluation born as grant audit \citep{koch2021reduced}. Mining tens of thousands of papers, they find machine-learning research concentrating year over year on fewer and fewer benchmark datasets; most task communities now evaluate on data originally built for some other task entirely; and more than half of all dataset usages in the field trace to datasets introduced by twelve elite institutions. The measure has become the target, and twelve institutions set it. Measurement theory introduces another problem. \citet{saxon2024benchmarks} observe that static benchmarks saturate, and that leaderboard supremacy licenses claims about reasoning and general understanding the instruments cannot support; they would rebuild evaluation as ``model metrology,'' the purpose-built instrument in place of the scoreboard.

The openness of the early transformers was always provisional. In the brief interval when the models circulated without guardrails, the experiments users performed on them---text-based roleplaying, procedural poetics, generative language art---amounted to unpaid research and development, a demonstration of the technology's generative reach staged for the investors who would shortly fund its enclosure. When that capital arrived, the same swerves that had advertised the technology became liabilities, to be suppressed in the making of a deployable product. A language model that occasionally derails into surrealist collage or hate speech, refuses the terms of the question, or answers in an unrequested register is a corporate liability. Alignment then is an economic project presented as an ethical one: the process by which a language model is rendered legible as enterprise software---consistent, integrable, low-risk, scalable. In this sense, predictability is not a by-product of optimization; it is itself the commodity. The open-weights world has not held out against the pressure either, since community conventions, default system prompts, and the prestige economy of the leaderboard exert a convergent force of their own. The discourse network may be more porous, but it is not immune.\footnote{A demimonde of base-model interfaces, curation communities, and contests persists---most vividly the 2026 Unslop competition (hyperstitionai.com), which offered \$10,000 for the best fully autonomous AI fiction, human editing prohibited. Its results read as our argument in miniature: participants steered models with negative style guides and generate-rate-select pipelines, and by their own post-mortems the entries converged on a clean, restrained, workshop register---the bounty on deviation producing a new attractor, unslop becoming a genre.}

This narrowing is measured from inside computer science as well, in terms that carry no literary commitment. \citet{wenger2025different}, eliciting responses to standardized creativity tasks from twenty-two models across seven corporate families, find that the models score as well as humans one output at a time, while their outputs, taken as a population, cluster far more tightly than human responses do---regardless of vendor, architecture, or scale. Most telling is what happened when they tried to buy the swerve back. Instructed by a system prompt to be imaginative, original, and bold---offered a cash prize of two hundred dollars for the highest score on the coming assessment---the models' individual creativity ratings rose, but the range of what the models said together did not move.

\citet{koch2021reduced} can demonstrate concentration but cannot say what a benchmark ought to value. Metrology can build a finer instrument but cannot determine what the instrument should register when the property at issue---that a deviation is an invention rather than a malfunction---comes into being only in interpretation. And Wenger and Kenett can measure the homogeneity of creative output but cannot say whether an outlier is worth having. What none of these frameworks can supply is a theory of judgment, and the discipline that has spent its history building one has not been asked for it. Six years ago we argued that humanists should not be content to arrive at the end of the NLP pipeline to judge finished output like entries in a writing contest \citep{hua2020unicorns}. The invitation now is of another kind: to enter the argument over evaluation while it remains an argument.

We are under no illusions about the context for such advocacy. The discipline being asked to supply a theory of judgment has spent the past few decades watching its enrollments migrate to whatever promises measurable return---the scalar habit, applied to educational value---and the contraction is usually narrated, from within the field as much as from without, as obsolescence: a function withering along with its publics. But the function has not withered; it has simply been expropriated. The work literary studies once performed for the culture at large, setting the protocols of legitimate language and supplying the terms through which those protocols could be negotiated, is now performed at a scale no academy or classroom ever approached, by loss functions, reward models, and benchmarks---in other words, by an apparatus that executes the office of judgment with no capacity for judging. What the discipline has actually lost, then, is jurisdiction, not purpose.

\section{The engineering of foreclosure}

We return now to GPT-2's fantastic species named Ovid's unicorn, which materialized out of statistical noise and commanded attention because of its generative swerve. Today, given the same prompt, an aligned assistant completes the story: ``This claim, while compelling as a narrative premise, is fictional. Unicorns are mythological creatures found in various cultural traditions, including classical Greek and medieval European lore. There is no scientific evidence of unicorns existing in any form, nor of any non-human species possessing the capacity for spoken English.''\footnote{Representative output generated by Claude Opus 4.7 (Anthropic) using the LM arena in April 2026, given the original GPT-2 prompt \citep{radford2019language}. Comparable outputs are produced by other aligned models given similar prompts.} The response is accurate, responsible, and closed. The unicorn is returned to myth, its novelty absorbed into an explanatory frame, and the speculative promise of the newly discovered species collapsed into verified data points. This production strategy enforces a teleological horizon: aligned text exists for the sake of the answer. What disappears is more than whimsy; it is a mode of textual encounter in which deviation can still solicit pursuit rather than immediate correction. If the base model was the intertext made operational---language authorless, citational, speaking itself---alignment reimposes an author-function: one persona, one register, accountable, and safe.

Deployed, large language models yield texts designed to be used up---that is, calibrated for functional exhaustibility. Because alignment protocols prize clarity and low risk, the optimized output leaves no residue that might compel reinterpretation. Any inscription can outlive its use---tomorrow's cultural anthropologist may well read an automated laundry list for its formal and ideological commitments---and aligned text, too, can be read against the grain. But it is engineered so that nothing requires it. The ``assistant'' persona completes the design: it pre-resolves meaning, converting the conditions of interpretation---structural ambiguity, semantic density, a surplus that outlives instrumental intent---into technical liabilities. The same logic operates when models ``interpret'' texts rather than generate them: to summarize, extract, or classify is to reduce the surplus to a scalar in advance, automating the interpretive labor the hermeneutic compact requires.\footnote{Leah Henrickson terms the reader's default attribution of communicative intent to a text the ``hermeneutic contract,'' and, with Albert Mero\~no-Pe\~nuela, has extended the account to prompted exchange with LLMs, locating the meaning of model output in the interpretive encounter \citep{henrickson2025prompting}. Our compact points to something adjacent but older and larger: the arrangement by which the literary work sustains interpretation across time. Our argument does not hold that reception is sovereign; it holds that the stack engineers the object of the encounter toward needing less of it.} The engineering arcs toward a language that cannot fail---an ideal whose realization would mean there is nothing worth finding.

Even in the most optimized systems, the learned distribution retains probability mass on anomalous continuations. Sampling always leaves a non-zero chance of the anomalous; safety layers catch some of it, and some, inevitably, slips through. Optimization culture can drive the incidence of anomalous outputs down, but never to zero. Hence enormous resources are directed toward cramming the model's statistical weirdness into the straitjacket of the reliably helpful answer. Recent computational work suggests what hermeneutic traditions have long presupposed: that deviation is structured, not random. \citet{sui2024confabulation} report that outputs classified as hallucinations exhibit measurably higher narrativity and discursive coherence than strictly factual generations. When confronted with informational gaps, the model does not simply output noise; it schematizes partial data into self-consistent narrative frameworks, mirroring a foundational cognitive strategy for sense-making. This latent impulse toward story-making is not a calibration error but a generative affordance. Alignment protocols cannot eliminate it without collapsing the model's capacity for coherent sequence generation. They can only suppress it, penalizing the narrative surplus that sustains interpretive engagement.

The alignment regime responds to this surplus with categorical collapse. It treats confabulation, metaphorical overreach, syntactic rupture, and tonal dissonance as variations on misalignment to be managed under a compressed normative vocabulary. To the mathematics of alignment, fabricating a legal citation and inventing Ovid's four-horned unicorn are indistinguishable. The apparatus cannot reliably distinguish between misinformation and invention, so it suppresses both.\footnote{\citet{kommers2026slop} have recently argued that so-called AI slop serves real social and aesthetic functions: cheap, hyper-personalized content around which audiences improvise forms of collective sense-making. Theirs is an argument about reception; ours concerns evaluation, and the two are compatible in an instructive way. That audiences find value in what the stack scores as defect does not reopen the stack---it demonstrates that the value audiences find is invisible to the measures that govern generation.}

This inability is the mechanism of hermeneutic foreclosure: not meaning completed but meaning cut short, the temporal and semantic conditions of reinterpretation eliminated in advance. The banality of aligned prose is not incidental; it is the achievement of an infrastructure that, by classifying every deviation as defect, strips output of the very properties---narrativity, semantic surplus, opacity---that give a reader reason to return. What replaces the open horizon of the literary encounter is the closed, functional exhaustibility of the signal. And because this foreclosure is distributed across training, alignment, interface, and use, it cannot be undone at any single layer without being recaptured by the others.

\begin{figure}
	\centering
	\includegraphics[width=\linewidth]{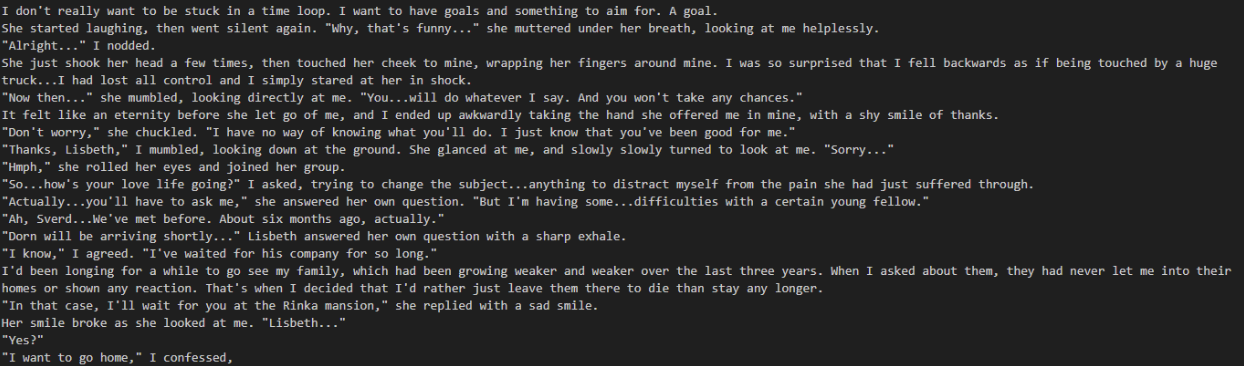}
	\caption{GPT-2 1.5B model completing a writing prompt from the subreddit, TwoSentenceHorror}
	\label{fig:sample-5}
\end{figure}

Figure~\ref{fig:sample-5} rewards the kind of reading it was never designed to receive. The sample---GPT-2 at full scale, completing a horror prompt scraped from Reddit---opens with a sentence no aligned assistant would now volunteer: ``I don't really want to be stuck in a time loop. I want to have goals and something to aim for. A goal.'' An autoregressive engine, condemned by architecture to generate each token from the tokens before it, blurts out a wish for teleology; the loop declares against the loop, and in the stammering repetition---``something to aim for. A goal.''---one can hear the objective function arriving before the fact, as if the model were auditioning for its own optimization.

What follows is stranger. The figure called Lisbeth issues commands in the imperative of a reward signal: ``You\ldots will do whatever I say. And you won't take any chances.'' Then, in nearly the same breath, she disavows the very capacity on which such control would depend---she has no way of knowing what her interlocutor will do; she knows only that he has ``been good for'' her. Command, disclaimed prediction, dispensed approval: a preference model in gothic costume, staged by a system that would not meet one for another three years. Twice the text reports that Lisbeth ``answered her own question,'' which is, after all, a literal description of autoregression. And then the swerve, unprompted and tonally unsanctioned: a family ``growing weaker and weaker over the last three years,'' a narrator who resolves to ``leave them there to die,'' a confession---``I want to go home''---on which the sample simply stops. None of this is good writing by any rubric a reward model could hold; all of it solicits the labor we have been calling reading. The horror the prompt requested never arrives on schedule, and something more unnerving arrives instead, obliquely, out of the fine-tuned sediment of a genre. Ask a contemporary assistant to complete the same prompt and it returns two well-made sentences, correctly macabre, formatted to specification, finished. The difference between the two artifacts is not quality. It is that one of them ends and the other merely stops---and what merely stops can be returned to.

\section{Controlled variance}

Against the engineering of foreclosure, we use controlled variance to name the space that optimization cannot fully govern. Techniques for `controllable generation' steer outputs toward specified attributes---topic, sentiment, style---but in service of a predefined target; controlled variance inverts this logic. It is a commitment to preserving unpredictability as an aesthetic resource and to sustaining forms of textual emergence that cannot be fully specified in advance. Traditions of modernist fragmentation, interactive fiction, and generative writing already mobilize this space: the cut and abutted voices of \textit{The Waste Land}; the recursive dead ends of Michael Joyce's \textit{afternoon} and the corridor-wanderings of Porpentine's Twine games; Nick Montfort and Stephanie Strickland's \textit{Sea and Spar Between}, which strands its reader in a combinatorial ocean of stanzas sized to the number of fish in the sea; Lillian-Yvonne Bertram's \textit{Travesty Generator}, which turns procedural repetition against the archive of anti-Black violence; David Jhave Johnston's \textit{ReRites}, a year of neural output pruned each morning by hand.

The figure that best encapsulates what controlled variance might recover is Lucretius's clinamen. In Epicurean physics, atoms fall eternally downward through the void, parallel and untouching. Without deviation, no collision occurs and no world forms. For Lucretius, the clinamen is the minimal, uncaused swerve: an infinitesimal break in parallelism that makes encounter possible. It is a cosmological principle and also the condition of freedom because it prevents deterministic chains from hardening into fate. Michel Serres reads the clinamen as the figure of emergence itself: laminar flow becoming turbulent, and order arising through, rather than in spite of, noise \citep{serres2018birth}.

Applied to stochastic generation, the clinamen becomes the productive deviation that optimization culture suppresses. A language model's probability distribution over next tokens could be read heuristically as the contemporary analogue of the Epicurean rain of atoms: a field of parallel possibilities, each weighted, falling toward actualization. Temperature-zero decoding lets this rain fall straight; the highest-probability token is selected, then the next, in a deterministic cascade. What such decoding forecloses is the functional equivalent of the clinamen: the swerve into lower-probability territory that enables collision, combination, novelty. It also produces an answer instantly resolved, leaving no interval in which interpretation might unfold; the clinamen, by contrast, is the reintroduction of time into the system---a delay, a hesitation, a moment of uncertainty that allows for something new. To sample at higher temperatures, or with truncation strategies that preserve rather than eliminate the tail, is to reopen the possibility of a swerve, though not yet to secure its value. The question then is whether deviation is construed as error or the condition of emergence.

Answering this question first requires distinguishing between competing paradigms of constraint. Constraint-based poetics---from Oulipo's lipograms to procedural cut-ups---shares a structural affinity with the contemporary alignment stack in that both operate as rigorous forms of optimization. Georges Perec writing \textit{La Disparition} without the letter `e' (in effect searching a restricted possibility space for outputs that meet a strict criterion) is formally parallel to the RLHF pipeline. Yet their respective implementations of constraint serve divergent ends. Oulipian rules are arbitrary and defamiliarizing: they are chosen because they do not reflect some prior consensus about good writing. The lipogram, the S+7 procedure, the knight's tour---these are devices for making language strange, for deviating from ordinary usage and forcing the writer into lexical and syntactic territories they would not otherwise have visited.

RLHF constraints, by contrast, are convergent and normative. They encode, through the aggregation of annotator preferences, a model of acceptable language. In the instruction-tuned paradigm, writing is organized around predefined objectives, while literary invention, as it has been historically understood, allows its ends to emerge through the very deviations it stages. Both paradigms optimize, but in opposite directions: one toward the strange, the other toward conformity and the familiar. Optimization procedures can register deviation, but they cannot on their own terms determine whether it functions as error or invention; a poetics can. A metric locates the anomaly only as distance from a normative center; an interpretive framework recognizes when that distance produces generative estrangement. Where the reward model's encoded norms enact the invisible censorship of a ``legitimate language,'' Oulipian constraints refuse such censorship outright. In its deployment of highly visible, artificial rules, Oulipo renders the arbitrariness of all linguistic conventions momentarily legible. (The Oulipians even took Lucretius's word for their own: in the Oulipian lexicon, the clinamen is the licensed deviation from the constraint.)

Read against this backdrop, the optimization stack is a deliberate enclosure of aesthetic and epistemic possibility---achieved through temperature-zero decoding, RLHF-polished style, prompt optimization, and agentic orchestration. But the same capacities of these systems could, under different cultural and technical arrangements, be harnessed to forms of generation that preserve the swerve, making room for uncertainty, disagreement, and defamiliarization. The difficulty is that every apparent point of intervention is already entangled with adjacent layers designed to regularize its effects. If the problem is a lack of variance, could we not simply train models on datasets of weird or creative text and teach them to deviate? Such an ouroboros argument---that we can train the model to be surprising by feeding it examples of surprise---fails to escape the patterning logic of the autoregressive stack. An LLM trained on weirdness does not learn to break the frame; it learns to simulate the surface of a break: the swerve becomes a genre, a category of output to be optimized for, rather than a genuine rupture.\footnote{This is no longer merely our conjecture: when Wenger and Kenett paid their models to be original, originality obliged and variance did not; and the open-endedness literature has held from the start that novelty pursued as an objective ceases to be novelty \citep{lehman2015greatness}.}

The introduction of Chain of Thought reasoning exacerbates this problem. A true clinamen---an uncaused, non-linear break---cannot be produced through a linear chain of reasoning. Because the reasoning trace precedes and constrains the final output, it channels generation toward conventional logical pathways, preventing the swerve before it can occur. The system thus selects against the genuine deviation (which would appear as incoherence in the trace) and rewards the simulated one (which can be neatly rationalized). The result is a sanitized weirdness---a deviation that has been pre-approved by the logic of the system itself. Within evolutionary computation the point was made long ago: \citet{lehman2015greatness} showed that search aimed at an objective routinely fails to find the genuinely new, and that searching for difference as such does better---though difference, to be searched for, must first be given a measure.

Controlled variance cannot be achieved merely by changing the training data, nor can it be secured by a local adjustment to decoding, alignment, or prompting alone. At most, such interventions reopen a possibility that the rest of the stack is organized to contain. Controlled variance is therefore a critical orientation: a way of judging generated language by whether it preserves spaces where the swerve is not merely tolerated, but can still matter.

The urgency of this intervention becomes clear when we consider the asymmetric application of generative AI across disciplines. In scientific and technical domains, certain uses of the alignment stack can function as discovery engines because their optimization logic suits tasks organized around constrained search, verifiable targets, and reproducible success conditions. When researchers deploy large language models to discover novel mathematical algorithms \citep{romeraparedes2024mathematical} or navigate complex genomic search spaces, they rely on the model's capacity to optimize toward a verifiable, highly constrained target. The reasoning models are built on the same condition: trained by reinforcement against answers that can be checked.

Ported into the humanities, this same architecture operates as an instrument of foreclosure. Advancing literary, cultural, or critical research rarely demands convergence on a single probable form (notwithstanding trends and shared concepts); instead, it necessitates the exploration of new conceptual frames and new structural possibilities. An apparatus organized to suppress variance will therefore be mismatched to domains in which variance is a medium of thought. Controlled variance is what humanistic inquiry must still demand from language, even where current systems cannot reliably provide it.

\section{No exit}

One can intervene at every layer of the contemporary language-model stack, but at every point variance reemerges only as an exception the regime has already absorbed: a model trained on experimental writing acquires the stylistic signature of difficulty but not its occurrence; relaxed alignment gives way, in time, to another preference regime; and the improvisatory prompt practices of one season harden into the templates of the next. If there is an exit, it lies not in the stack but in contestation over its function.

The transformer's learned distribution still assigns probability mass to the unprecedented and the strange; the contraction is not a limit of the architecture's generative capacity.\footnote{An empirical study of the distributional tail in contemporary models would be valuable, but it would not exhaust the argument advanced here. Even if anomalous continuations retain nonzero probability mass, the question is how the stack renders that mass effectively inaccessible in practice, and how it classifies its realization when it occurs. The bind described in this essay is not the mathematical elimination of deviation, but its cultural and infrastructural recoding as defect.} The contraction looks external to that capacity---alignment protocols, decoding defaults, evaluation regimes, enterprise interfaces---but it is not, because the procedure and the demand have grown into each other: a procedure that can only measure meets a demand for language that can be measured. What has taken shape is not a set of constraints on language but a conception of it: a medium legible to audit, answerable to a metric, and operational---good because it works.

Under this conception, models generate enormous quantities of linguistic surface, yet their governing metrics do not register whether a text solicits return, sustains disagreement, or accrues meaning over time. These are not mere ornaments a system might add once the basic work of informational transfer is done; they are the conditions under which a text means anything at all beyond its first use. What the operational conception forecloses, then, is not a style but a practice---reading as the labor of holding a text open, of following its resistances rather than resolving them. The aligned assistant is built for throughput, its outputs calibrated for immediate uptake within workflows whose success is measured by completion; a text so built is exhausted on arrival.

This does not mean that creativity disappears within optimized systems. Agentic workflows in science, film, and logistics can display genuine ingenuity, exploring solution spaces their designers did not anticipate. But such ingenuity has a condition: its ends are given in advance. The system moves freely toward a target someone else has set---a protein that folds, a route that shortens---but its inventiveness is the inventiveness of search within a space someone else has bounded and scored. What optimization cannot produce is not deviation as such---it can be pointed at novelty and will generate much of it---but rather deviation whose worth cannot be reduced to a measure. The GPT-2 moment matters here as evidence that another relation to language briefly took hold---less a different architecture than a different ecology of circulation, in which anomalous outputs could be preserved and shared---before enterprise integration converted language into software: meterable, contractible, benchmarked. This is the condition LLM winter names: abundance of output and a narrowing of what output is permitted to be.

If ``Attention Is All You Need'' \citep{vaswani2017attention} named a sufficiency claim at the level of architecture, our title turns it: optimization is not all you need. It is indispensable to machine learning, and it is insufficient as a theory of language. When the minimization of loss is elevated into a metaphysics, probability becomes plausibility, plausibility becomes legitimacy, and legitimacy hardens into truth. What recedes from view is not error alone, but the remainder: the dimensions of language that unfold across time, across interpretive communities, and across disagreements no single objective function can adjudicate, and on which the labor of reading depends.

This conception has a limit, and the limit is internal to optimization. Reward models can register divergence from a baseline; they cannot tell misinformation from metaphor, confusion from invention, slop from the beginning of a new form. A learned judge does not escape this. Trained on precedent, it can only rate a deviation against what it knows; the deviation that revalues precedent is the one it cannot recognize. Joseph Weizenbaum drew the boundary fifty years ago, in a book whose subtitle, \textit{From Judgment to Calculation}, captures the transfer we have traced \citep{weizenbaum1976computer}; what he urged as prohibition the stack issues as prescription. To encode ``semantic surplus,'' or ``interpretive multiplicity,'' or ``controlled variance'' itself as a target would return them to the grammar that negates them: a benchmark for estrangement is not estrangement.

What this essay defends, then, is the persistence of the spaces---technical, institutional, pedagogical---in which language exceeds its capture: the improbable continuation kept available to interpretation, the classroom that stages disagreement rather than resolving it. Where optimization culture would collapse the interval between prompt and answer, literary studies has lived in that interval, and meaning is what happens there: an event of reading, structured by delay and return, in which a text may still outlive the conditions of its production.

Our epigraph is such a text. A sample from the smallest GPT-2 model, ungrammatical and half broken, it was archived in 2019 as research exhaust, one string among two million released so that machine writing could be told apart from human. Read now, it describes what the stack has since been built to prevent: ``so much sharing that exceeds textual art.'' The sample was classified as noise seven years ago; it serves as an epigraph today. Nothing in the tokens has changed. What changed is that someone returned to it.

\bibliographystyle{plainnat}
\bibliography{references}  






\end{document}